\documentclass[a4paper,10pt]{llncs}

\usepackage{algorithmicx}
\usepackage{algpseudocode}
\usepackage[utf8]{inputenc}
\usepackage[american]{babel}
\usepackage[pdftex]{graphicx}
\usepackage{color}
\usepackage[colorlinks=false,pdfborder={0 0 0}]{hyperref} 
\usepackage{xspace}
\usepackage{amsmath}
\usepackage{ifthen}
\usepackage{subfig}
\usepackage{cleveref}
\usepackage{morefloats}
\usepackage{balance}
\usepackage{amssymb}
\usepackage[autolanguage]{numprint}
\usepackage{multirow}
\usepackage{multicol}
\usepackage[colorinlistoftodos,prependcaption,textsize=small,disable]{todonotes}
\usepackage{ctable}
\usepackage{color}
\usepackage{paralist}
\usepackage{tabularx}
\usepackage{footnote}
\usepackage[normalem]{ulem}
\usepackage{csquotes}
\usepackage{paralist}
\usepackage{booktabs}
\usepackage[normalem]{ulem}
\usepackage[inline]{enumitem}
\usepackage[sortcites,
backend=biber
]{biblatex}
\usepackage{tikz}
\usetikzlibrary{shapes,arrows}

\AtEveryBibitem{%
   \clearfield{urlday}%
   \clearfield{urlmonth}%
   \clearfield{urlyear}%
   \clearfield{isbn}%
   \clearfield{pages}
   \clearfield{url}
   \clearfield{editor}
   \clearfield{doi}
   \clearfield{issn}
   \clearlist{publisher}
   \clearfield{address}
   \clearlist{location}
}

\newcommand{\aho}[1]{\todo[author=aho,color=pink]{#1}}
\newcommand{\tni}[1]{\todo[author=tni,color=yellow]{#1}}
\newcommand{\mbe}[1]{\todo[author=mbe,color=green]{#1}}
\newcommand{\chp}[1]{\todo[author=chp,color=blue!20]{#1}}

\newcommand{\dmir}{Data Mining and Information Retrieval Group}

\newcommand{\eg}{e.g.,\xspace}
\newcommand{\ie}{i.e.,\xspace}

\newcommand{\wiki}{Wikipedia\xspace}
\newcommand{\bibs}{BibSonomy\xspace}
\newcommand{\delicious}{Delicious\xspace}
\newcommand{\wikinav}{WikiNav\xspace}
\newcommand{\wikiglove}{WikiGloVe\xspace}

\newcommand{\glove}{GloVe\xspace}

\newcommand{\wsbig}{WS-353\xspace} 
\newcommand{\bibeval}{Bib100\xspace}

\newcommand{\men}{MEN\xspace}

\newcommand{\incr}{$\nearrow$}
\newcommand{\decr}{$\searrow$}

\newcommand{\furl}[1]{\footnote{\scriptsize\url{#1}}}

\newcommand{\para}[1]{\noindent\textbf{#1}}

\makeatletter \@ifundefined{BibTeX}
   {\def\BibTeX{{\rmfamily B\kern-.05em%
    \textsc{i\kern-.025em b}\kern-.08em%
    T\kern-.1667em\lower.7ex\hbox{E}\kern-.125emX}}}{}
\makeatother

\newcommand{\egfurl}[1]{\footnote{e.g.~\url{#1}}}

\title{Learning Semantic Relatedness from Human Feedback Using Metric Learning}
\author{Thomas Niebler\inst{1*}
\and Martin Becker\inst{1}
\and Christian Pölitz\inst{1}
\and Andreas Hotho\inst{1,2}}
\institute{\dmir, University of Würzburg (Germany)\\
\email{\{niebler, becker, poelitz, hotho\}@informatik.uni-wuerzburg.de}
\and
L3S Research Center Hanover (Germany)\\
}

\begin{document}

\maketitle

\begin{abstract}
Assessing the degree of semantic relatedness between words is an important
task with a variety of semantic applications, such as ontology learning for the
Semantic Web, semantic search or query expansion.
To accomplish this in an automated fashion, many relatedness measures have been
proposed.
However, most of these metrics only encode information contained in the
underlying corpus and thus do not directly model human intuition.
To solve this, we propose to utilize a metric learning approach to improve
existing semantic relatedness measures by learning from additional information,
such as explicit human feedback.
For this, we argue to use word embeddings instead of traditional
high-dimensional vector representations in order to leverage their semantic
density and to reduce computational cost.
We rigorously test our approach on several domains including tagging
data as well as publicly available embeddings based on \wiki texts and
navigation.
Human feedback about semantic relatedness for learning and evaluation is
extracted from publicly available datasets such as \men or \wsbig.
We find that our method can significantly improve
semantic relatedness measures by learning from additional
information, such as explicit human feedback.
For tagging data, we are the first to generate and study embeddings. 
Our results are of special interest for ontology and recommendation engineers,
but also for any other researchers and practitioners of Semantic Web
techniques.
\end{abstract}

\section{Introduction}
\label{sec:introduction}
Automatically assessing semantic relatedness as perceived by humans is a task
with many applications related to semantic technologies, such as ontology
learning for the Semantic Web, semantic search or query expansion.
Recent work has shown that semantic relatedness between words can successfully
be extracted from a wide range of sources, such as
tagging data~\cite{cattuto2008semantic,markines2009evaluating},
\wiki article texts~\cite{gabrilovich2007computing,radinsky2011computing}
or \wiki navigation~\cite{niebler2015extracting,singer2013computing}.
In particular, such approaches usually encode semantic information of words
in continuous \emph{word vectors}~\cite{turney2010frequency}.
The semantic relatedness between two words can then be measured by the cosine
similarity of their corresponding word vectors.
While the above-cited methods come close to human intuition of semantic
relatedness, they are only able to encode information contained in the
underlying corpus.
Thus, they do not explicitly represent the actual notion of semantic relatedness
as expected and employed by humans.
The natural way to solve this problem is to incorporate additional
information, such as explicit human feedback, in order to account for the
deviations of the respective semantic relatedness measure from human intuition.
Furthermore, such feedback could be helpful to adapt semantic relatedness
measures to specific domains and tasks or to personalize them in order to
improve recommendation approaches or retrieval methods for search engines.

\para{Problem Setting and Approach.}
Consequently, this work addresses the issue of incorporating additional
information such as explicit human feedback about semantic relatedness 
into relatedness measures operating on vector representations of words.
To this end, we apply a metric learning approach where we encode the additional
information in the form of constraints.
This manipulates the original relatedness measure and ultimately
yields a better fit with human intuition.

%

There are many domains from which semantic relatedness has been extracted.
Thus, we aim to propose a universally applicable approach.
To illustrate the flexibility of this method, we apply it to 
\wiki article \emph{texts},
\emph{navigational traces} on the \wiki page network, 
and \emph{tagging data}.
To represent words in these application domains we use word embeddings, \ie
low-dimensional, dense vector representations, which reduce the computational
complexity of our metric learning approach and have been shown
to outperform high-dimensional representations in
measuring semantic relatedness~\cite{baroni2014count}.
While word embedding approaches have been applied to several \wiki domains
(texts or
navigation)~\cite{pennington2014glove,levy2014linguistic,dallmann2016extracting},
we are, to the best of our knowledge, the first to derive tag embeddings and
study the relationship between their dimensionality and their semantic
expressiveness.

Independent of the domain, we show that our approach can significantly improve
the quality of the given semantic relatedness measures.
As mentioned earlier we confirm this on different domains by learning from and
evaluating on a variety of well known semantic relatedness datasets generated
from human intuition.
In this context, we study the influence of the amount of information used for
learning and investigate if the improved semantic relatedness measures
generalize between different human intuition datasets.

\para{Contribution.} Our contribution is twofold:
First, we introduce the metric learning setting (with relative constraints) to
  the domain of semantic relatedness and --- by exploiting human feedback ---
  show that it is possible to use metric learning to improve semantic
  relatedness measures to better fit human intuition.
  This also opens a new connection for the field of semantic relatedness
  research to a popular field in machine learning and can lead to another
  fruitful combination of both.
  We explicitly show how to adopt the metric learning scenario for our
  relatedness learning setting. 
Secondly, we are the first to generate and study embeddings from
  tagging data, which allows for effectively and efficiently performing metric
  learning.

Overall, our work describes a way to improve semantic relatedness measures
based on additional semantic information, \eg explicit human feedback.
This enables us to increase the fit of these measures to human intuition
significantly and even introduce user-specific information into the
corresponding semantics.
Our results are of special interest for ontology and search engineers,
but also for any other practitioners of Semantic Web techniques.

\para{Structure.} The rest of this paper is structured as follows: in
\Cref{sec:methodology}, we introduce our approach of learning semantic
relatedness using metric learning.
In \Cref{sec:expsetup}, we describe the conducted experiments and present the
results, which we discuss in \Cref{sec:discussion}.
\Cref{sec:relwork} gives an overview of other work related to this paper.
Finally, \Cref{sec:conclusion} concludes this work and gives directions for
future research.

\section{Metric Learning to Learn Semantic Relatedness}
\label{sec:methodology}
%
In this section, we first formulate the goal of learning semantic relatedness in
terms of metric learning.
We then argue that the notions of distance and semantic relatedness are
equivalent when restricting the setting to a transformed unit sphere.
Finally, we introduce our method to formulate human feedback as
constraints for the metric learning algorithm we employ.
\Cref{fig:metriclearning_pipeline} shows a sketch of the steps that we apply in
our approach to learn semantic relatedness.

\tikzstyle{data} = [rectangle, draw, text centered, text
width=6em]
\tikzstyle{optional} = [rectangle, draw, fill=gray!30, draw=gray!80, text
centered, text width=6em, text=gray!90]
\tikzstyle{operation} = [draw, rectangle, text width=5em, text
centered]

\tikzstyle{line} = [draw, -latex']

\begin{figure}[b!]
	\centering
	\begin{tikzpicture} 
		\node [data] (rawdata) at (0, 1.5) {Raw Data (\emph{\scriptsize Tagging Data,
		\wiki, etc.})};
		
		\node [operation] (embeddings) at (2.5, 1.5) {Word Embeddings
		(\emph{\scriptsize \glove, Word2Vec, etc.})};
		
		\node [operation] (metriclearning) at (5, 1.5) {\textbf{Metric Learning
		(\emph{LSML})}};
		\node [data] (hids) at (10, 1.5) {Explicit User Feedback ($\mathcal{H}$)};
		\node [operation] (constraints) at (7.5, 1.5) {Constraint Generation
		($\mathcal{C}$)};
		
		\node [data] (relatednessmeasure) at (5, 0) {\textbf{Relatedness
		Measure $\mathbf{\cos_M}$}};
		
		\node [optional] (evaluation) at (8, 0) {Evaluation (Spearman)}; 
		
		\draw[->, blue!50, very thick, rounded corners=5pt] (rawdata) to (embeddings);
		\draw[->, blue!50, very thick] (embeddings) to (metriclearning);
		\draw[->, blue!50, very thick] (hids) to (constraints);
		\draw[->, blue!50, very thick] (constraints) to (metriclearning);
		\draw[->, blue!50, very thick] (metriclearning) to (relatednessmeasure);
		\draw[->, gray!50, very thick] (relatednessmeasure) to (evaluation);
		\draw[->, gray!50, very thick] (hids) |- (evaluation);
	
	
	\end{tikzpicture}
	\caption{The pipeline of our approach.
	We preprocess raw data in order to create word embeddings, which serve as
	vector input for the metric learning algorithm.
	Simultaneously, we transform a portion of the user feedback information
	$\mathcal{H}$ to relatedness constraints $\mathcal{C}$ for the metric learning
	algorithm.
	The output of the algorithm is a relatedness measure $\cos_M$, characterized by
	a symmetric, positive-definite matrix $M$.
	Later, this matrix together with a portion of the explicit user feedback is
	used for the evaluation, which is explained in~\Cref{sec:expsetup_evaluation}.}
	\label{fig:metriclearning_pipeline}
\end{figure}
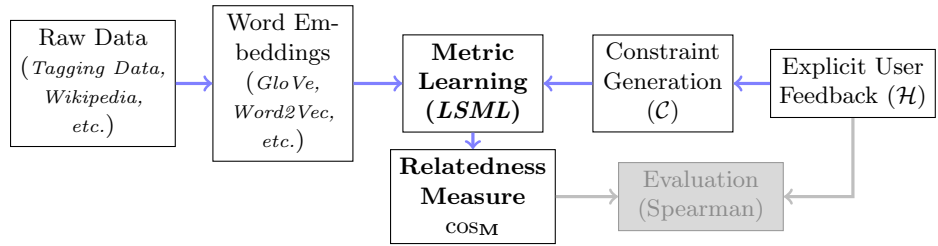

\para{Learning Semantic Relatedness in Terms of Metric Learning.}
To learn a metric, standard metric learning algorithms parameterize the
Mahalanobis distance,
$d_M(x,y) = \sqrt{(x-y)^T M(x-y)}$, by finding a (symmetric, positive definite)
matrix $M$.
To this end, most algorithms expect a set of constraints $\mathcal{C}$.
In this work, we apply the LSML algorithm~\cite{liu2012metric}, which learns
from relative distance constraints of the form
$
\mathcal{C} := \left\{(x, x', y, y') : d(x, x') < d(y, y')\right\}.
$

However, instead of learning a distance $d_M(x, y):
\mathbb{R}^n \times \mathbb{R}^n \rightarrow \left[0; \infty\right)$ we want to
learn a \textit{semantic relatedness} measure $rel_M(x, y):
\mathbb{R}^n \times \mathbb{R}^n \rightarrow \left[-1; 1\right]$.
Accordingly, instead of formulating constraints based on a distance $d$, they
are based on a relatedness measure $rel$, \ie in our case the cosine ($cos$).
Overall, this amounts to learning a symmetric,
positive definite matrix $M$ such that the parametrized cosine measure
$\cos_M(x, y) := x^TMy \cdot (\|x\|_M \|y\|_M) ^{-1}$, where $\|x\|_M :=
\sqrt{x^TMx}$,
suffices all constraints $\mathcal{C}$ which are derived from human
intuition on semantic relatedness (instead of distance).
In the following, we show that --- when restricting the setting to a transformed
unit sphere --- learning a distance measure and a semantic relatedness measure is
equivalent when specifying the constraints correctly.

\para{Equivalence of Distance and Relatedness for Metric Learning.}
In the following, we first show the equivalence of distance and
relatedness (as measured by the Mahalanobis metric and the parameterized
cosine measure, respectively) on the transformed unit sphere $\mathbb{S}_M^{n -
1} := \left\{ x \in \mathbb{R}^n | \|x\|_M = 1 \right\}$.
This allows us to formulate constraints in a way to learn semantic relatedness.

For all vectors on the transformed unit sphere $x,y \in \mathbb{S}^{n-1}_M$
parametrized by $M$, the Mahalanobis distance metric and the parameterized cosine measure
$cos_M(x,y)$ can be expressed as each other.
That is, since $x^TMx = 1$ and because of $M$ being symmetric ($x^TMy = y^TMx$),
it holds that:
	$d_M^2(x, y) = 2\cdot(1 - \cos_M(x, y))\label{eq:dcos}$.
This is in line with the intuition that if two words are more closely related,
the distance of their vector representations is lower and vice versa.
Thus, we can directly apply the metric learning approach to learning
semantic relatedness measures on word vectors, if the constraints are correctly
specified. 
To this end, we use the fact that the previous equation trivially implies:
\begin{eqnarray*}
	d_M (x, x') <  d_M(y, y') &
	\Leftrightarrow & \cos_M(x, x') > \cos_M(y, y')
\end{eqnarray*}
Thus, we can formulate the constraints based on a relatedness notion $rel$
instead of a distance $d$ by specifying $\mathcal{C} := \left\{(x, x', y, y') :
rel(x, x') > rel(y, y')\right\}$, \ie the comparison operator is inverted.

\para{Obtaining Relatedness Constraints from Human Feedback.}
To obtain suitable constraints to train the metric learning algorithm for
learning semantic relatedness, we propose to exploit human intuition
datasets.
Such datasets contain word pairs together with human-assigned relatedness scores
(see also \Cref{sec:evaluation_datasets}) which can be interpreted as explicit
human feedback.
Formally, such datasets can be expressed as a set $\mathcal{H} :=
\left\{ (w_i, w'_{i}, r_i) \right\}$, where $w_i$ and
$w'_i$ are words and $r_i$ is the human-assigned score which describes an
intuitive notion of the degree of relatedness between the two corresponding
words.
As such relatedness scores are commonly collected in a crowdsourcing
task~\cite{finkelstein2001placing,bruni2014multimodal}, they thus represent
explicit human feedback.
To obtain constraints in the form $(x, x', y, y')$ to use in the metric
learning algorithm, one can simply combine all pairs of examples $(w_i, w'_i,
r_i), (w_j, w'_j, r_j) \in \mathcal{H}$ with $r_i < r_j$ and thus receive a set
of relatedness constraints $\mathcal{C} = \left\{(w_i, w'_i, w_j,
w'_j)\right\}$.

Since we want to emphasize the importance of some constraints in the
optimization step of the algorithm, we place higher weights on those constraints
to make sure that they are fulfilled.
The LSML algorithm allows to assign weights to all constraints.
In order to put a high emphasis on a constraint with one very unrelated pair of
words, \eg $(w_i, w'_i, r_i)$, and another very related pair of words, \eg
$(w_j, w'_j, r_j)$ with $r_i \ll r_j$, we can define the weight of this
constraint according to the difference of the respective human relatedness
scores of the corresponding word pairs.
The extended constraints can then be written as 
$\mathcal{C}_{weighted} := \left\{(w_i, w'_i, w_j, w'_j, r_j - r_i)\right\}$.
In the remainder of this work, we always employ this kind of weighted
constraints.

\section{Datasets}
\label{sec:datasets}
In this work, use two different kinds of datasets to evaluate our metric
learning approach to integrate user feedback to relatedness measures.
That is, \emph{domain datasets} which provide a set of word vectors representing
the words to calculate semantic relatedness for, and \emph{human intuition
datasets} (HIDs) which we employ to learn semantic relatedness and to test our
results.
In the following we first describe two domain datasets containing tagging
data from which we later derive tag embeddings.
Then we review two domain datasets based on \wiki which come with
pre-trained word vectors.
Finally, we introduce all human intuition datasets containing human-assigned
scores of similarities to word pairs.

\subsection{Tagging Datasets to Derive Word Embeddings}
In our work, we study datasets of two public social tagging systems.
We use data from \bibs, which has a more academic audience.
The second dataset is a subset of the \delicious social tagging system, where
the audience is focused on design and technical topics.

Each dataset is restricted to the top 10k tags to reduce noise.
Additionally, we only considered those tags from users who have tagged at
least 5 resources and only those resources which have been used at least 10 times.
We also removed all invalid tags, \eg containing whitespaces or unreadable
symbols.

\para{\bibs.}
The social tagging system \bibs provides users with the possibility to collect
bookmarks (links to websites) or references to scientific publications and
annotate them with tags~\cite{benz2010social}.
We use a freely available dump of \bibs, covering all tagging data from 2006
till the end of 2015.\furl{http://www.kde.cs.uni-kassel.de/bibsonomy/dumps/}
After filtering, it contains \numprint{10000} distinct tags, which were assigned
by \numprint{3270} users to \numprint{49654} resources in \numprint{630955}
assignments.

\para{\delicious.}
Like \bibs, \delicious is a social tagging system, where users can share their
bookmarks and annotate them with tags. We use a freely available dataset from
2011~\cite{zubiaga2013harnessing}.\furl{http://www.zubiaga.org/datasets/socialbm0311/}
\delicious has been one of the biggest adopters of the tagging paradigm and due
to its audience, contains tags about design and technical topics.
After filtering, the \delicious dataset contains \numprint{10000} tags, which
were assigned by \numprint{1685506} users to \numprint{11486080} resources in
\numprint{626690002} assignments.

\subsection{Pre-trained Embedding Datasets Based on \wiki}
In order to demonstrate the applicability of our approach on any kind of word
embeddings, we also use two publicly available datasets of pre-trained vectors.
Both are related to \wiki, which has been shown time after time to yield high
quality semantic
content~\cite{dallmann2016extracting,gabrilovich2007computing,niebler2015extracting,radinsky2011computing,singer2013computing}.

\para{\wikiglove.} 
The authors of the \glove embedding algorithm~\cite{pennington2014glove}
trained several datasets of vector embeddings on various text data and made them
publicly available.\furl{https://nlp.stanford.edu/projects/glove/}
Because it has been demonstrated several times that the textual content of \wiki
articles can be exploited to calculate semantic
relatedness~\cite{gabrilovich2007computing,radinsky2011computing},
we use the vectors based on \wiki as a reference for word embeddings generated
from natural language.
This dataset consists of \numprint{400000} vectors with dimension 100.

\para{\wikinav.} 
Wulczyn published a set of word embeddings generated from navigation data on the
\wiki webpage~\cite{wulczyn2016wikipedia} using
Word2Vec~\cite{mikolov2013distributed}.
Word2Vec was originally intended to be applied on natural language text,
though it can also be applied on navigational
paths~\cite{dallmann2016extracting}.
While technically the generated embeddings represent pages in \wiki, most pages
also describe a specific concept and can thus be used interchangeably.
It has been shown that exploiting human navigational paths as a source of
semantic relatedness yields meaningful
results~\cite{niebler2015extracting,singer2013computing,west2009wikispeedia}.
The dataset at hand consists of \numprint{1828514} vectors with 100 dimensions.
The vector embeddings have been created from all navigation data in the month of
January 2017 and are publicly
available.\furl{https://figshare.com/articles/Wikipedia_Vectors/3146878}



\subsection{Human Intuition Datasets (HIDs)}
\label{sec:evaluation_datasets}
\begin{table}[t!]
	\centering
	\caption{Overview of all Human Intution Datasets (HIDs). For each HID, we
	give the number of word pairs, the number of unique words and the number of
	judgments per word pair. Also, for each embedding dataset, we give the number
	of matchable pairs, where both words are present in the dataset's vocabulary.}
	\begin{tabularx}{\textwidth}{X|XX|X|X|X|X}
		\toprule
		dataset &  pairs &  words &\multicolumn{4}{c}{matches}
		\\
				&		&		   & \bibs 
		& \delicious 
		& \wikinav 
		& \wikiglove 
		\\
		\midrule
		\bibeval &    100 &    122 &   100 &         94 & 42 &         98 \\
		   \men  &   3000 &    751 &   465 &       1376 &     1227   &       3000 \\
		  \wsbig &    353 &    437 &   158 &        202 &      173   &        353 \\
		\bottomrule
	\end{tabularx}
	
	\label{tbl:hid_overview}
\end{table}

As a gold standard for semantic relatedness as it is perceived by humans, we use
several datasets with human-generated relatedness scores for word pairs, so
called human intuition datasets (HIDs).
They will provide training as well as test data.
In the following, we will describe all used HIDs briefly.
\Cref{tbl:hid_overview} gives an overview of the dataset sizes and the overlap as well as the Spearman
correlation for the matchable pairs for all embedding datasets.

\para{\wsbig.}
The
WordSimilarity-353\furl{http://www.cs.technion.ac.il/~gabr/resources/data/wordsim353/wordsim353.html}
dataset consists of \numprint{353} pairs of English words and
names~\cite{finkelstein2001placing}.
Each pair was assigned a relatedness value between \numprint{0.0} (no relation)
and \numprint{10.0} (identical meaning) by \numprint{16} raters, denoting the
assumed common sense semantic relatedness between two words.
Finally, the total rating per pair was calculated as the mean of each of the
\numprint{16} users' ratings.
This way, \wsbig provides a valuable evaluation base for comparing our concept
relatedness scores to an established human generated and validated collection
of word pairs.

\para{\men.}
The MEN Test Collection~\cite{bruni2014multimodal} contains
\numprint{3000} word pairs together with human-assigned similarity judgments,
obtained by crowdsourcing using Amazon Mechanical
Turk\furl{http://clic.cimec.unitn.it/~elia.bruni/MEN}.
Contrary to \wsbig, the similarity judgments are relative rather than absolute.
Raters were given two pairs of words at a time and were asked to choose the pair
of words was more similar.
The score of the chosen pair, \ie the pair of words that was more similar, was
then increased by one.
Each pair was rated \numprint{50} times, which leads to a score between
\numprint{0} and \numprint{50} for each pair.

\para{\bibeval.}
The \bibeval dataset has been created in order to provide a more fitting
vocabulary for the more research and computer science oriented tagging data
that we investigate.\furl{http://www.dmir.org/datasets/bib100}
It consists of \numprint{122} words from the top \numprint{3000} words of the
\bibs dataset and combined them into \numprint{100} word pairs, which
subsequently were judged 26 times each for semantic relatedness using
crowdsourced scores between \numprint{0} (no similarity) and \numprint{10}
(full similarity).\tni{Das hat keine Zitation. Soll ich hier einfach das
Evaluating Semantics-Paper auf Arxiv laden?}

%

\section{Experimental Setup and Results}
\label{sec:expsetup}
In this section, we perform several sets of experiments in order to demonstrate
the usefulness of metric learning for learning semantic relatedness.
First, we describe how we evaluate the quality of a learned semantic relatedness
measure.
Then we study the procedure of generating word embeddings from tagging data and
perform a qualitative evaluation.
Finally, we 
train several metrics on a range of domains considering different amounts of user
feedback,
investigate whether it is possible to transport trained semantic knowledge across
different collections of user feedback and
finally assess the robustness of the learned semantic relatedness measures.
We publish our code to enable reproducibility of our
experiments.\furl{http://dmir.org/semmele}

\subsection{Evaluating the Quality of Semantic Relatedness Measures}
\label{sec:expsetup_evaluation}
Most of the time, the quality of semantic relatedness measures is assessed by
how well it fits human
intuition~\cite{gabrilovich2007computing,mikolov2013distributed,singer2013computing}.
Human intuition is collected in Human Intuition Datasets (HID) as introduced in
\Cref{sec:evaluation_datasets}.
The most widely-used method to evaluate semantic relatedness on such
datasets is the Spearman rank correlation coefficient
which compares the ranking of word pairs given by a HID with the ranking implied
by the semantic relatedness measure.
%
%
While there exist other evaluation approaches like analogy
matching~\cite{levy2014linguistic,mikolov2013distributed} or concept
categorization~\cite{baroni2014count}, they do not fit our setting, because we
exclusively want to improve measuring relatedness.


\subsection{Word Embeddings from Tagging Data}
We evaluate our approach on several domains. 
This includes semantic relatedness extracted from tagging data.
However, vector representations of words extracted from tagging data are
traditionally high-dimensional~\cite{cattuto2008semantic,markines2009evaluating},
making metric learning in this domain infeasible due to a more than quadratic
runtime with regard to the number of vector dimensions \cite{liu2012metric}.
Thus, similar to our \wiki examples we employ the notion of (low-dimensional)
word embeddings which have been shown to outperform their
high-dimensional counterparts in terms of correlation with human intuition of
semantic relatedness~\cite{baroni2014count}.
This can also be confirmed when using tagging data as input (see
\Cref{tbl:hid_correlations}, cf. $\rho_{high}$ vs. $\rho_{emb}$).
In this section, we justify our choice of using \glove to embed words based on
tagging data and study the influence of dimensionality on the respective
semantic content.

\begin{table}[t!]
	\centering
	\caption{Spearman correlation scores for both the high-dimensional
	representation ($\rho_{high}$) and word embeddings ($\rho_{emb}$) of both
	tagging datasets.
	It can be seen that the word embeddings encode semantic relations which are
	more in line with human judgment than the high-dimensional representations.}
	\begin{tabularx}{.6\textwidth}{X|XX|XX}
	\toprule
	datasets	& \multicolumn{2}{c}{\bibs}	   & \multicolumn{2}{c}{\delicious}\\
				& $\rho_{high}$	& $\rho_{emb}$ & $\rho_{high}$	& $\rho_{emb}$ \\
	\midrule
	\bibeval	& 0.621			& 0.726		   & 0.640			& 0.675\\
	\men		& 0.436			& 0.483		   & 0.581			& 0.752\\
	\wsbig		& 0.395			& 0.575		   & 0.454			& 0.690\\
	\bottomrule
	\end{tabularx}
	\label{tbl:hid_correlations}
\end{table}
\para{Choosing an Embedding Algorithm.}
In this work, we apply the \glove algorithm~\cite{pennington2014glove}, which
learns word embeddings from a word co-occurrence matrix.
Other candidates are the well-known Word2Vec approach 
by~\cite{mikolov2013distributed} and the LINE
algorithm~\cite{tang2015largescale}.
However, Word2Vec relies on the meaningfulness of the sequential order of
words which is not available from tagging data.
LINE --- which learns node embeddings preserving the first and second order
neighborhood of the nodes in the graph --- is more applicable.
However, we found that it has a tendency to perform even worse than standard
high dimensional representation for calculating semantic relatedness.
Thus, overall we only report results on \glove, since it was directly
applicable to tagging data and showed the best results in our experiments.

\para{Embedding Dimension.}
One decision to make when generating word embeddings is choosing their
dimension:
We want the number of dimensions to be small to reduce the complexity of
the metric learning approach, but it needs to be large enough to encode the
necessary semantic information.
In order to find a good embedding representation of the tags, we experimented
with the dimension of the generated vector embeddings on the \delicious and
\bibs tagging data measuring semantic relatedness using the standard
cosine measure.
Due to the internal random initialization of \glove, we ran the vector 
embedding generation process 10 times for each number of dimensions in order to
study the corresponding standard deviations.

\begin{figure}[t!]
	\centering
	\subfloat[BibSonomy]{\includegraphics[width=.48\textwidth]{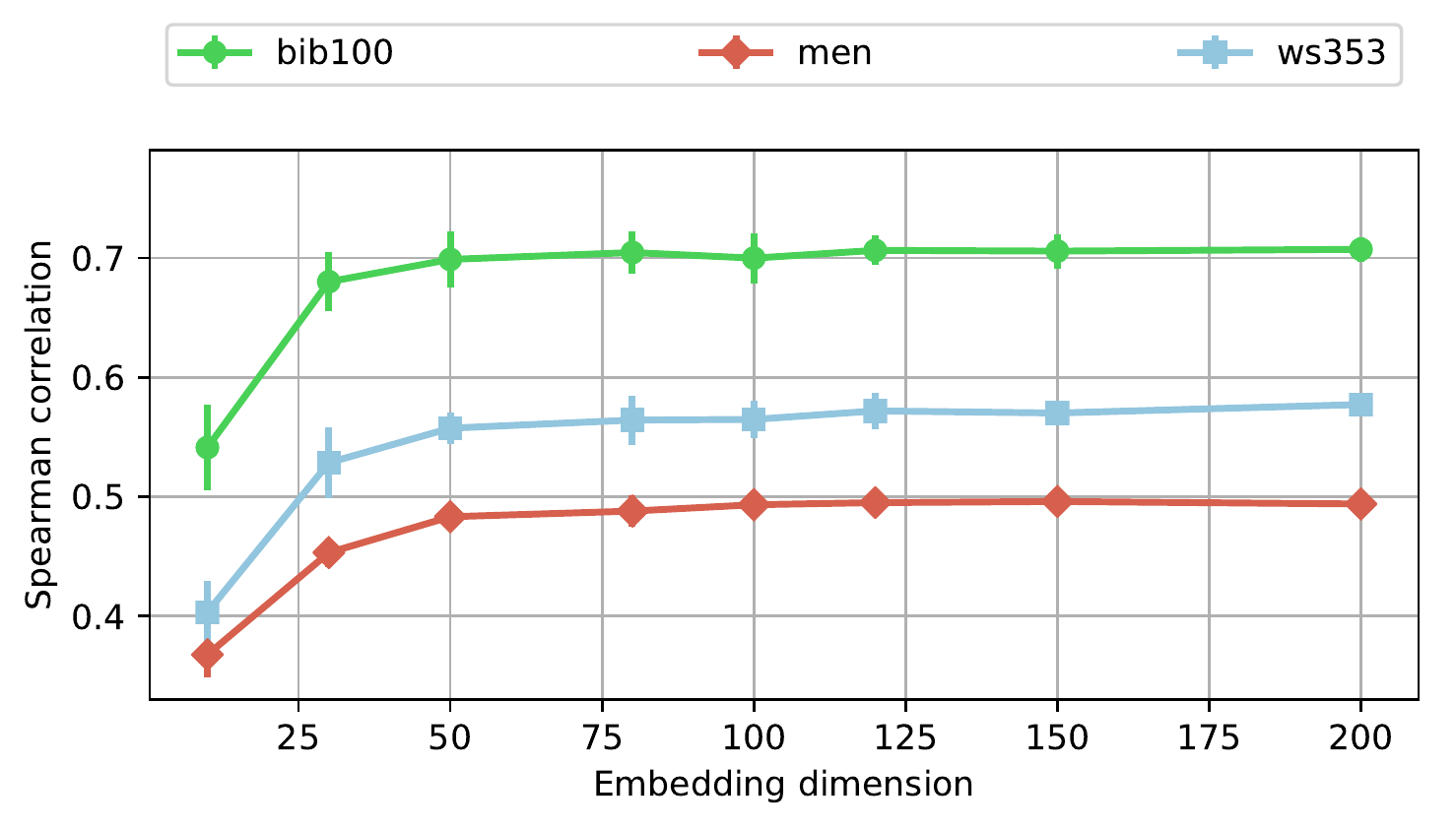}\label{fig:embeddings_bibsonomy_variance}}
	\subfloat[Delicious]{\includegraphics[width=.48\textwidth]{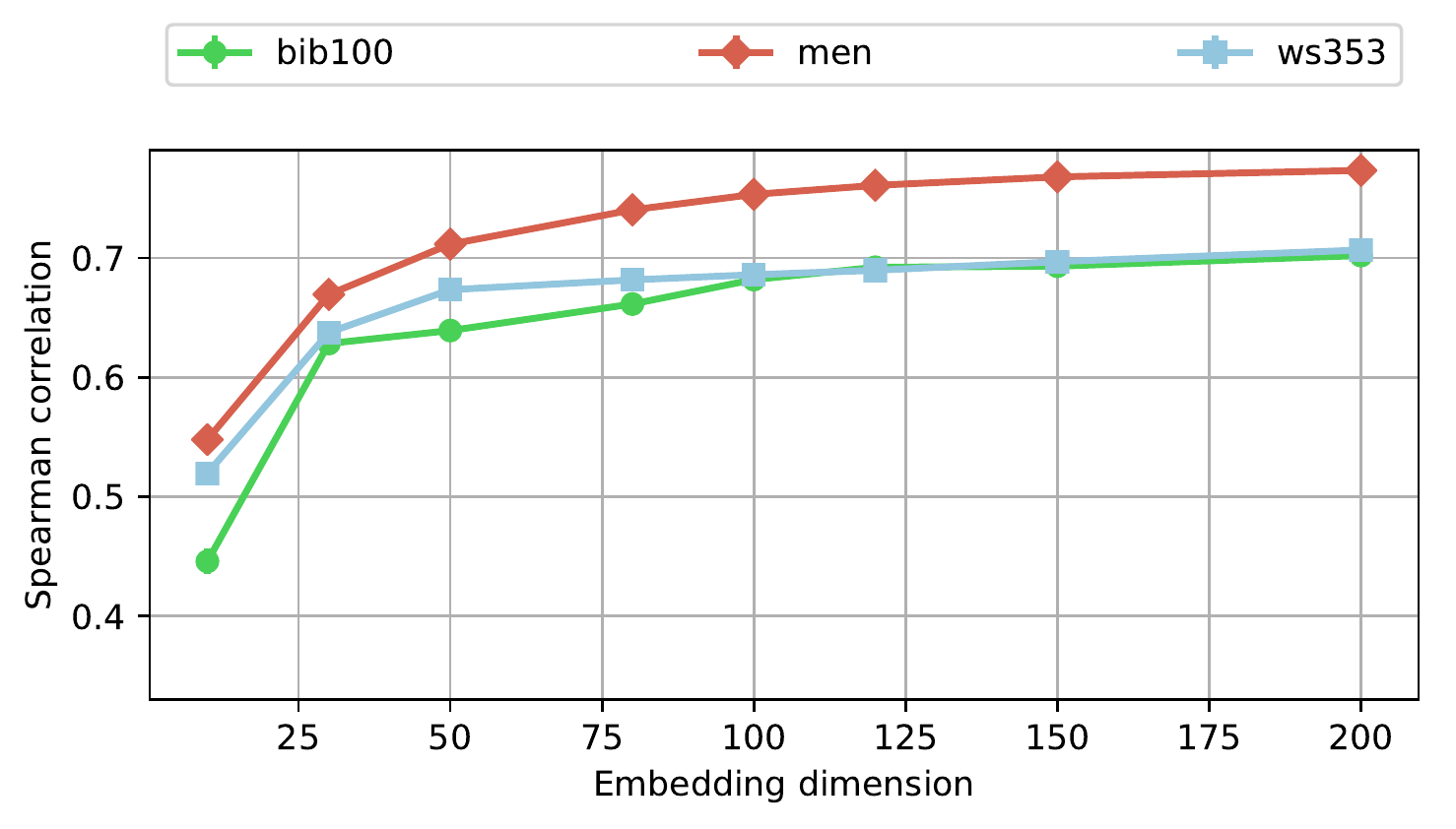}\label{fig:embeddings_delicious_variance}}
	
	\caption{Impact of the vector dimension and random initialization of the
	embedding algorithm on the evaluation result on different HIDs across several
	vector dimension settings.
	The error bars show the standard deviation, the dots depict the mean of the
	evaluation results across 10 runs with the same parameter settings.
	}
	\label{fig:embeddings_dimension_variance}
\end{figure}

The results for both experiments are shown in
\Cref{fig:embeddings_dimension_variance}.
%
For \bibs the influence of the random initialization of \glove on the semantic
content of the vectors decreases with increasing dimensionality, as indicated by
the error bars.
For the larger and denser \delicious dataset, there is less room for the random
initialization to influence the results.
This explains the hardly visible standard deviations.
With regard to the semantic content of the embeddings, the increase in semantic
quality settles around a dimension of 100 for \bibs.
For \delicious, adding more dimensions keeps increasing the semantic
content.
However, the 100 mark also signifies a drastic inhibition of the growth-rate. 
Thus, considering the quadratic training complexity of the LSML algorithm in
terms of vector space dimension, 
we decided to perform all of the following experiments on tagging data with
100-dimensional embeddings.


\subsection{Integrating Different Levels of User Intentions}
\label{sec:randomsampling}
In this section we investigate how the amount of human feedback used for
training influences the quality of the learned semantic relatedness measure.
To this end, we evaluate various training set sizes extracted from the HIDs on
the different embedding sets (pre-trained or extracted by \glove, cf.,
\Cref{sec:datasets}).

For each HID, we first randomly sampled 20\% of all matchable pairs as test
sets. 
We gathered 5 such test datasets.
For each test set, we sample training sets of different sizes (10\% - 100\%)
on which we train a metric each.
This metric is evaluated on the 20\% of previously sampled test data.
We repeat sampling training sets and learning 25 times.
Then, for each training set size, we take the mean and the standard deviation
over all experiments.
As a baseline, we also report the Spearman correlation using the
pure cosine measure on the test datasets. 
\begin{figure}[t!]
	\centering
	\subfloat[\bibs]{\includegraphics[width=.48\textwidth]{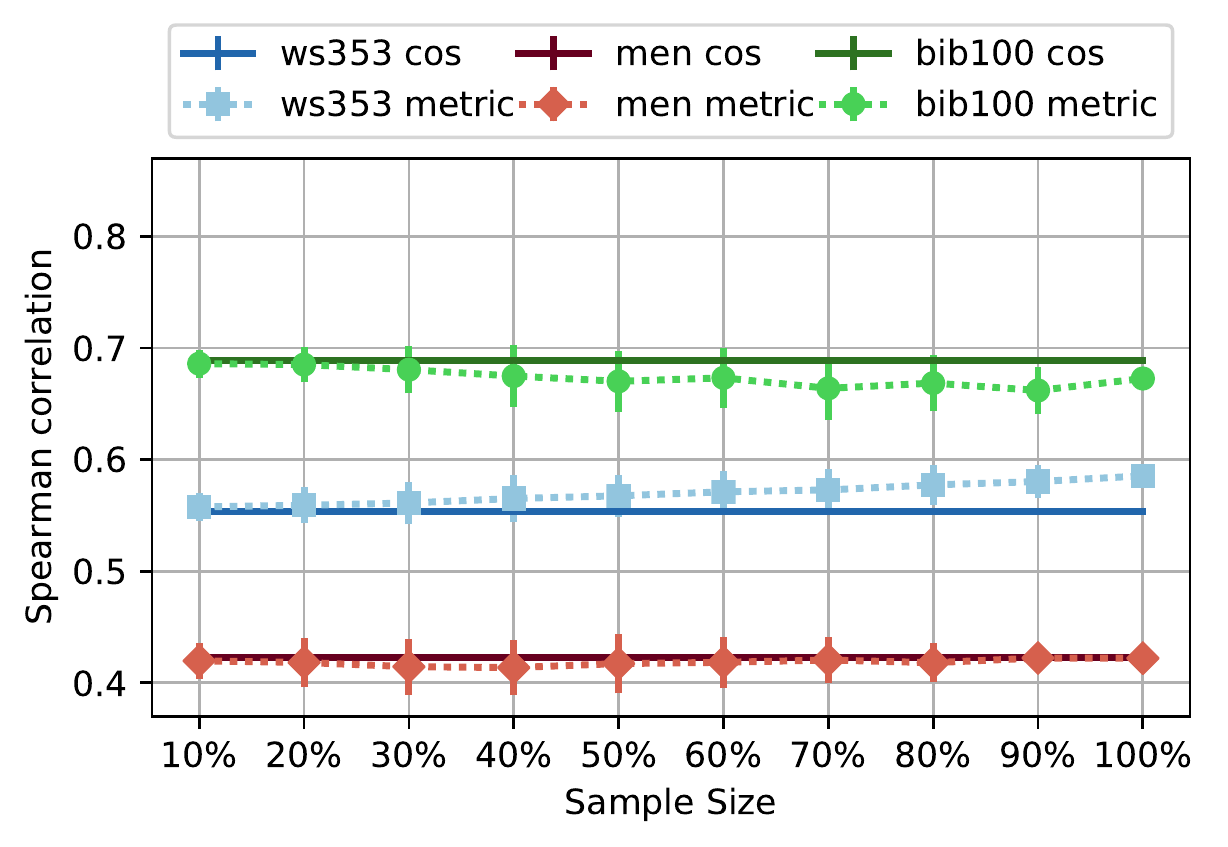}\label{fig:results_bibsonomy_randomsampling}}
	\subfloat[\delicious]{\includegraphics[width=.48\textwidth]{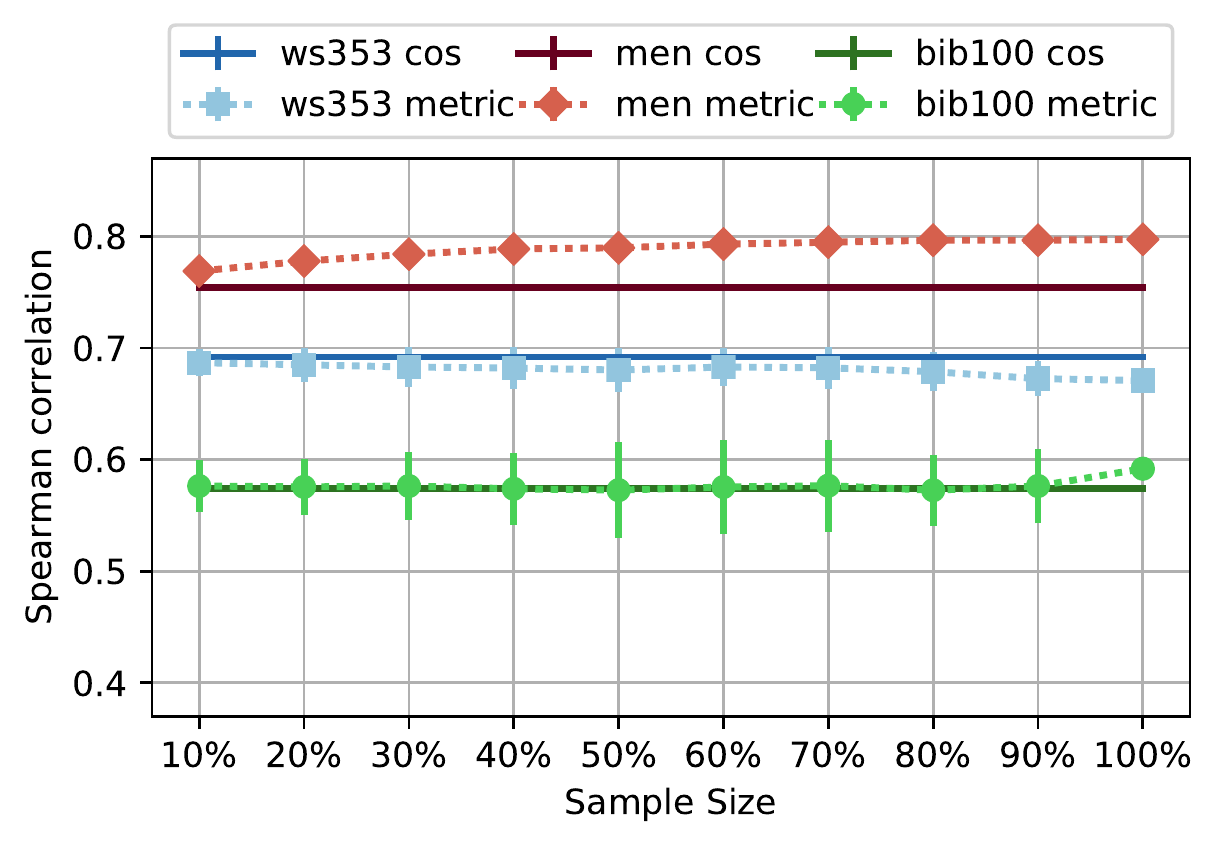}\label{fig:results_delicious_randomsampling}}
	
	\subfloat[\wikiglove]{\includegraphics[width=.48\textwidth]{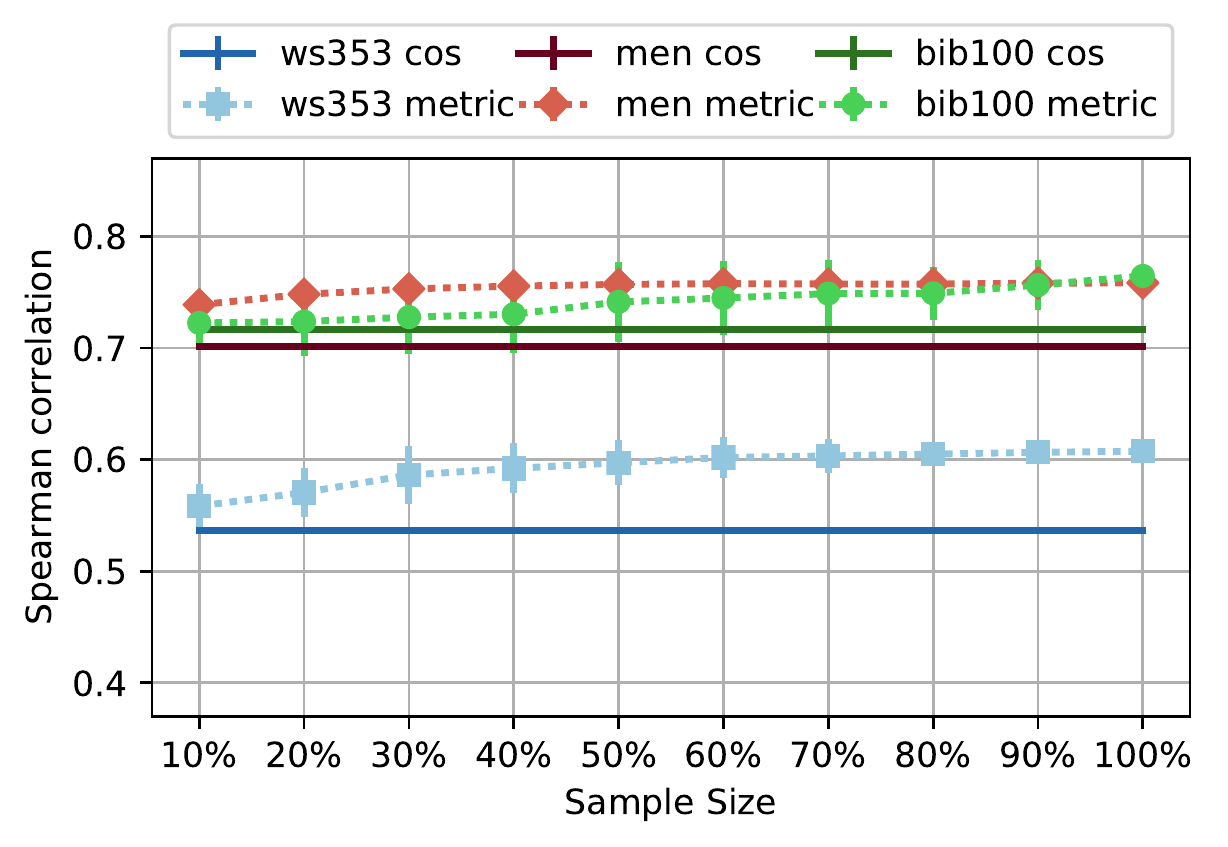}\label{fig:results_wikiglove_randomsampling}}
	\subfloat[\wikinav]{\includegraphics[width=.48\textwidth]{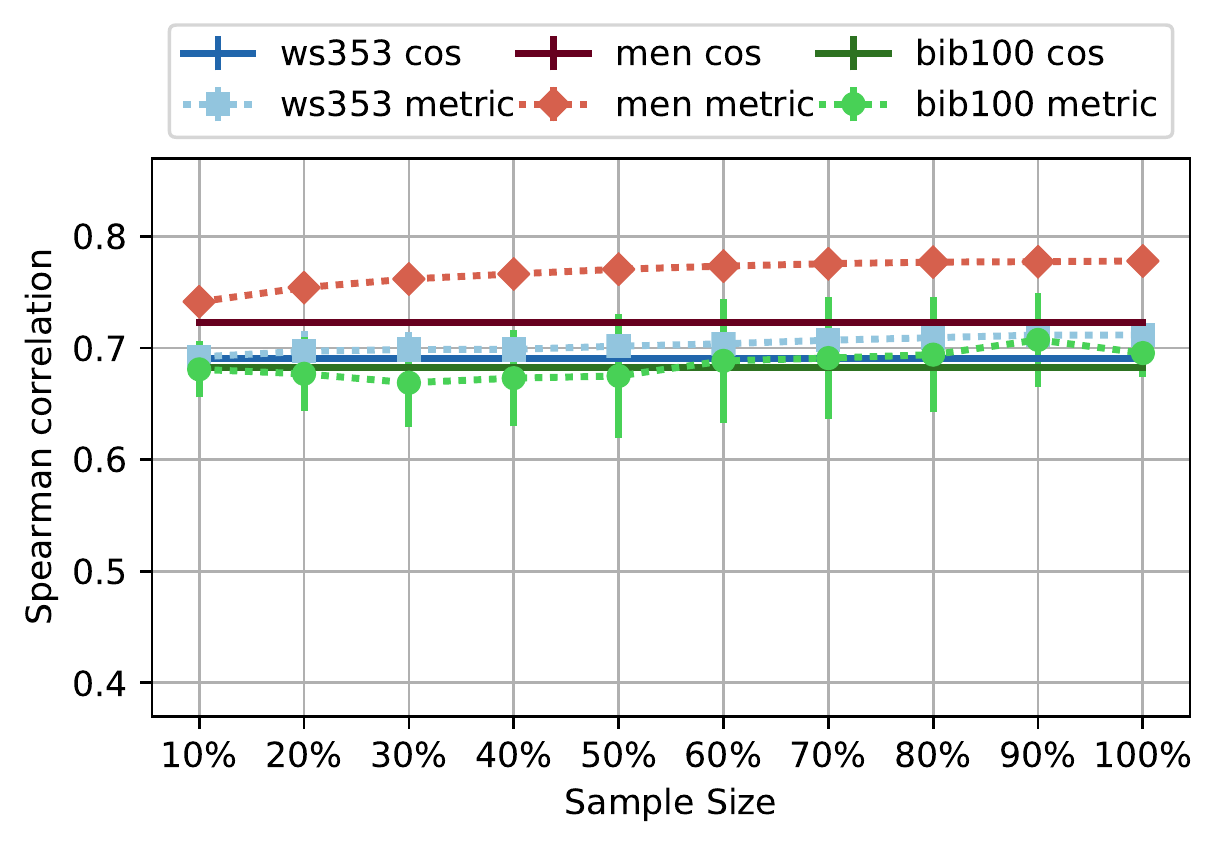}\label{fig:results_wikivectors_randomsampling}}
	
	\caption{Results on different levels of user feedback training data. 
	The dashed lines show the mean Spearman correlations on the test data, using
	the trained metrics, together with the standard deviation of the results.
	The continuous lines report the Spearman values when applying the
	standard cosine on the test data. 
	}
	\label{fig:results_randomsampling}
\end{figure}
\mbe{Ich würde generell über ``learning the notion of semantic relatedness in
MEN'' reden, statt zu sagen ``MEN is suited for learning semantic relatedness''}
\Cref{fig:results_randomsampling} shows that we can indeed inject user feedback
information about semantic relatedness into our relatedness measure.
The results indicate that we can best learn semantic relatedness with the
human intuition encoded in the \men dataset.
This is consistent across all datasets except the \bibs embeddings.
While it is also sometimes possible to use the knowledge from the \wsbig dataset
to improve our semantic relatedness measure, it does not improve results as
much as knowledge from other human intuition datasets.
On the \delicious embeddings, it even decreases performance to learn from
\wsbig knowledge.
Surprisingly, while the \bibeval dataset yields the best results on
the \bibs embeddings using the plain cosine measure, we cannot exploit the
contained knowledge enough to learn semantic relatedness from it.
\aho{Verstehe ich nicht. Was heißt das? Es bedeutet doch auch, dass es
eine Beziehung zwischen vectorraum und HID gibt, oder?}
\tni{Da hab ich Andreas' Kommentar nicht verstanden.}
Also, across all four embedding sets, the \bibeval dataset shows the biggest
variance of results, while the standard deviation of the \men dataset results
are tightly bounded.
We can generally observe that using vectors from \wikiglove for training seems
to be beneficial for our approach, as we can always improve the fit of our
measure to human intuition significantly, regardless of the choice of training
data.

\subsection{Transporting User Intentions}
The previous experiments showed that the integration of a dedicated HID into a
relatedness measure results in higher agreement of the measure with human
intuition.
Now, in order to transfer different user intentions across different settings,
we trained metrics on one complete HID and evaluated them on a different
HID.
For example, training was done using all \wsbig relations but the metric was
evaluated on the \men HID.
By this we evaluate if the learned knowledge generalizes from one notion of
semantic relatedness (represented by a specific HID) to another.

Results are given in \Cref{tbl:results_crossevaluation}.
For each line, its header defines the dataset on which the metric was trained,
while the column header is the dataset on which the trained metric was then
evaluated.
In each cell, the first value denotes the Spearman correlation of the cosine
measure with the human relatedness scores in the evaluation dataset.
The second value is the Spearman correlation of the relatedness scores
calculated from the trained metric with the human relatedness scores in the
evaluation dataset.
Depending on whether the trained metric increased or decreased correlation with
human intuition, we depict upwards or downwards arrows.
\begin{table}[t!]
	\centering
	\caption{Results for user intention transport experiments. We trained a
	metric on all word pairs from the dataset given at the start of each line and evaluated
	them on the dataset given in the column header. The first value is the
	Spearman correlation for the cosine measure on the evaluation dataset, the
	second value is the Spearman value for the trained metric. The arrow denotes
	if we could transfer relevant information from one dataset to another or not.}
	\subfloat[\bibs]{\scriptsize
		\begin{tabular}{l|c c c}
		\toprule
				& \men				& \wsbig			& \bibeval\\
		\midrule
		\men	& -					& 0.576\incr 0.591	& 0.726\decr 0.673\\
		\wsbig	& 0.484\decr 0.475	& - 				& 0.726\decr 0.687\\
		\bibeval& 0.484\decr 0.462	& 0.576\decr 0.557	& - 	\\
		\bottomrule
		\end{tabular}
		\label{tbl:hid_results_bibs}
	}
	\subfloat[\delicious]{\scriptsize
		\begin{tabular}{l|ccc}
		\toprule
				& \men				& \wsbig			& \bibeval\\
		\midrule
		\men	& -					& 0.690\decr 0.682	& 0.676\decr 0.644\\
		\wsbig	& 0.752\incr 0.766	& - 				& 0.676\decr 0.652\\
		\bibeval& 0.752\incr 0.772	& 0.690\decr 0.679	& - 	\\
		\bottomrule
		\end{tabular}
		\label{tbl:hid_results_del}
	}

	\subfloat[\wikiglove]{\scriptsize
		\begin{tabular}{l|ccc}
		\toprule
				& \men				& \wsbig			& \bibeval\\
		\midrule
		\men	& -					& 0.533\incr 0.604	& 0.658\incr 0.726\\
		\wsbig	& 0.693\incr 0.729	& - 				& 0.658\incr 0.700\\
		\bibeval& 0.693\incr 0.727	& 0.533\incr 0.601	& - 	\\
		\bottomrule
		\end{tabular}
		\label{tbl:hid_results_wiglo}
	}
	\subfloat[\wikinav]{\scriptsize
		\begin{tabular}{l|ccc}
		\toprule
				& \men				& \wsbig			& \bibeval\\
		\midrule
		\men	& -					& 0.729\incr 0.751	& 0.738\decr 0.737\\
		\wsbig	& 0.709\decr 0.703	& - 				& 0.738\decr 0.715\\
		\bibeval& 0.709\incr 0.715	& 0.729\decr 0.718	& - 	\\
		\bottomrule
		\end{tabular}
		\label{tbl:hid_results_winav}
	}
	\label{tbl:results_crossevaluation}
\end{table}

From \Cref{tbl:results_crossevaluation}, we can see some interesting results:
Generally, training a metric on \bibs embeddings almost always yields bad
transfer results.
The only improvement using \bibs embeddings is on the \wsbig dataset, when
evaluating the metric trained on \men.
The \delicious embeddings are only useful for improving relatedness scores when
the trained metric is evaluated on the \men dataset.
The most interesting part of these results is that, using \wikiglove embeddings,
we can always increase correlation with human intuition by a notable margin.
However, on the \wikinav embeddings, the results differ only by smaller margins.
The only exception here is a notable improvement on the Spearman correlation
value of \wsbig, when using a metric trained on \men data.
\chp{was sagt uns das jetzt insgesammt? warum ist es nur bei \wikiglove besser?}
Overall, it is generally possible to transfer knowledge from one HID to another.
However, this is highly dependent on the underlying word representations.


\subsection{Robustness of the Learned Semantic Relatedness Measure}
Here we inject wrong semantic relatedness information into
our learning process.
The goal is to show that 
i) wrong ratings do not collapse the relatedness measures, which ultimately
makes our approach robust for different users with different intuitions of relatedness, and 
ii) that the promising results of the previous experiments are indeed caused by 
the successful injection of user feedback.


The setup is the same as with the random sampling experiment
(\Cref{sec:randomsampling}), 
except that we randomly reassign the relatedness scores of the training pairs.
This way, we evaluate on valid human intuition, but learn from false
information.
\begin{figure}[t!]
	\centering
	\subfloat[\bibs]{\includegraphics[width=.48\textwidth]{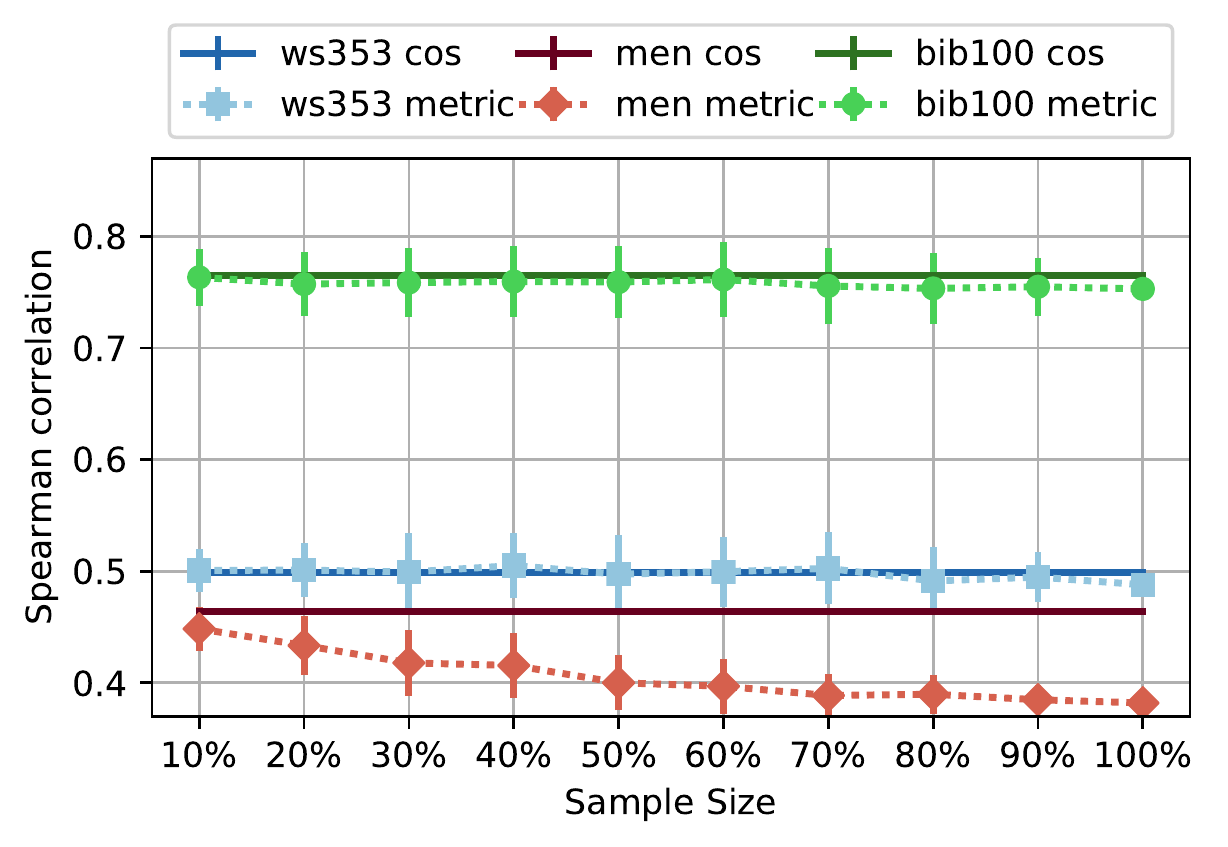}\label{fig:results_bibsonomy_randomsampling}}
	\subfloat[\delicious]{\includegraphics[width=.48\textwidth]{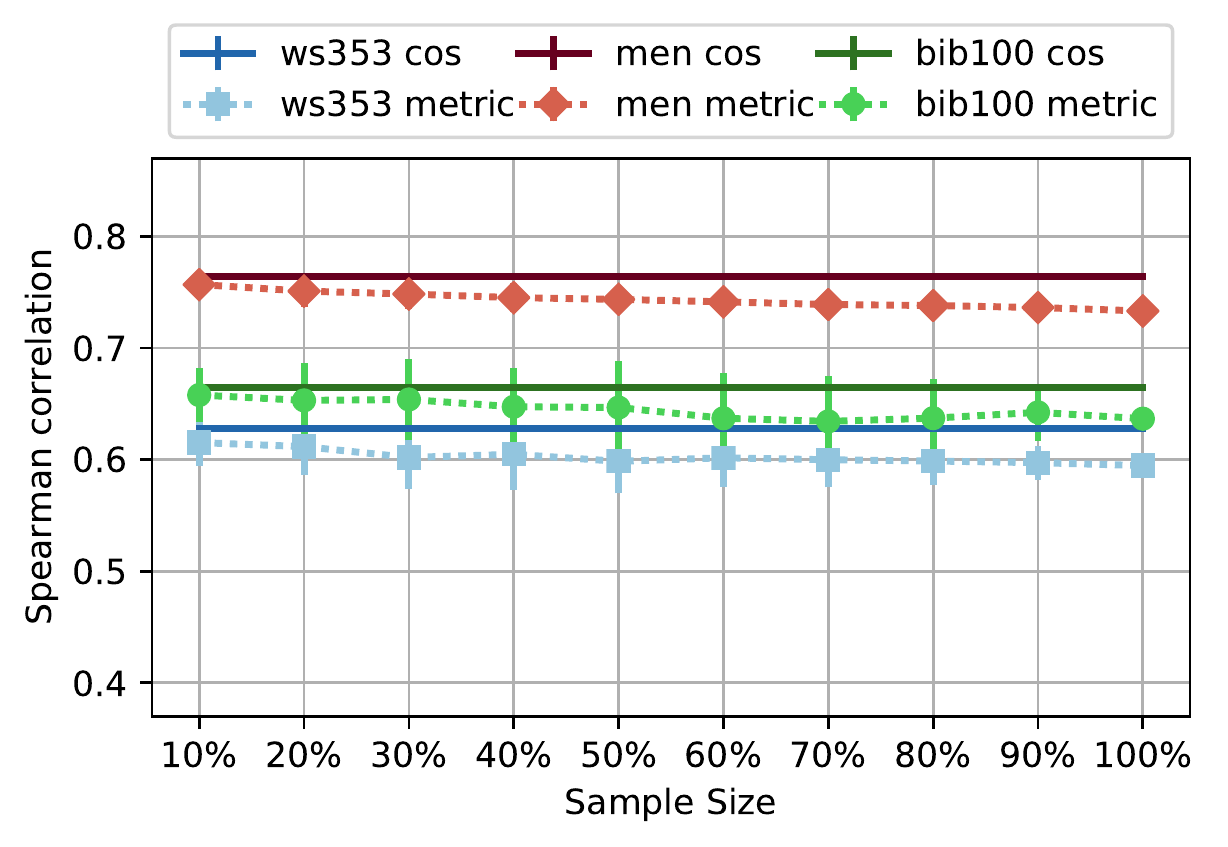}\label{fig:results_delicious_randomsampling}}
	
	\subfloat[\wikiglove]{\includegraphics[width=.48\textwidth]{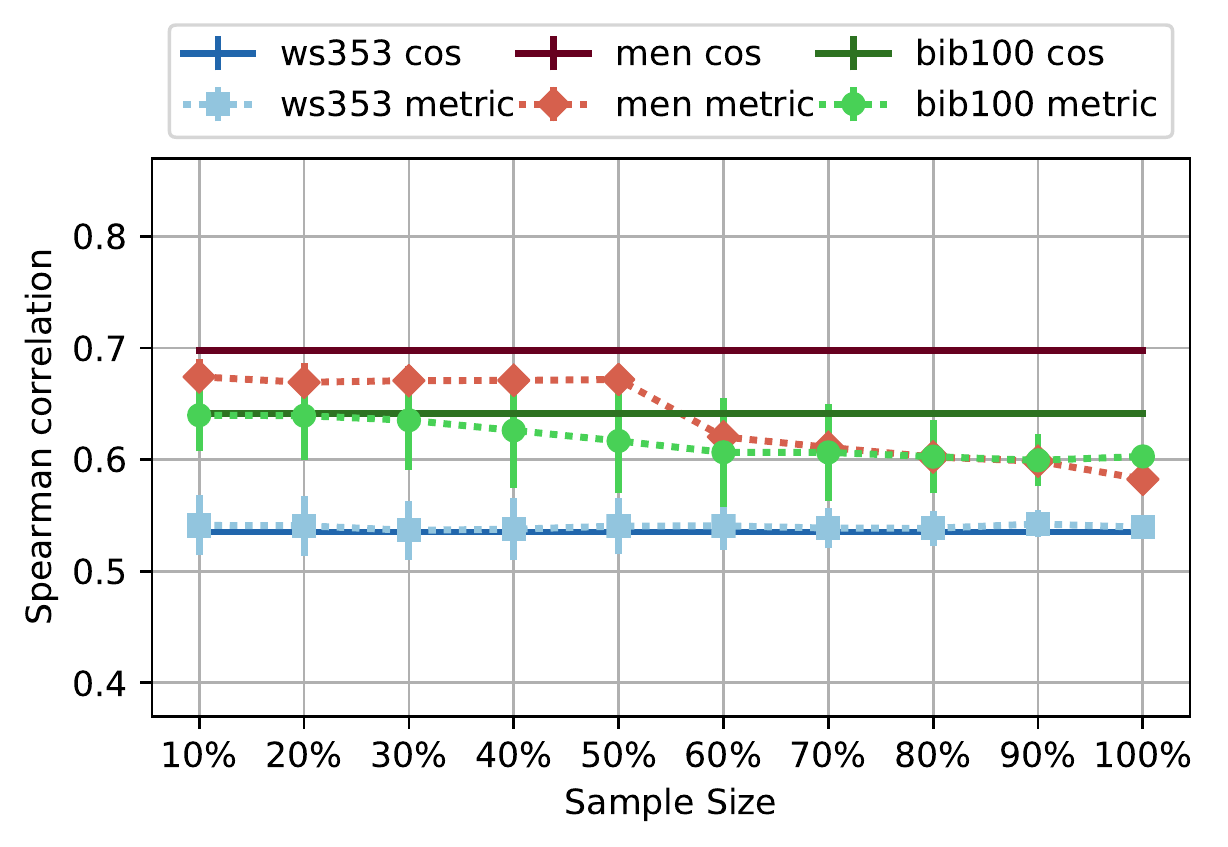}\label{fig:results_wikiglove_randomsampling}}
	\subfloat[\wikinav]{\includegraphics[width=.48\textwidth]{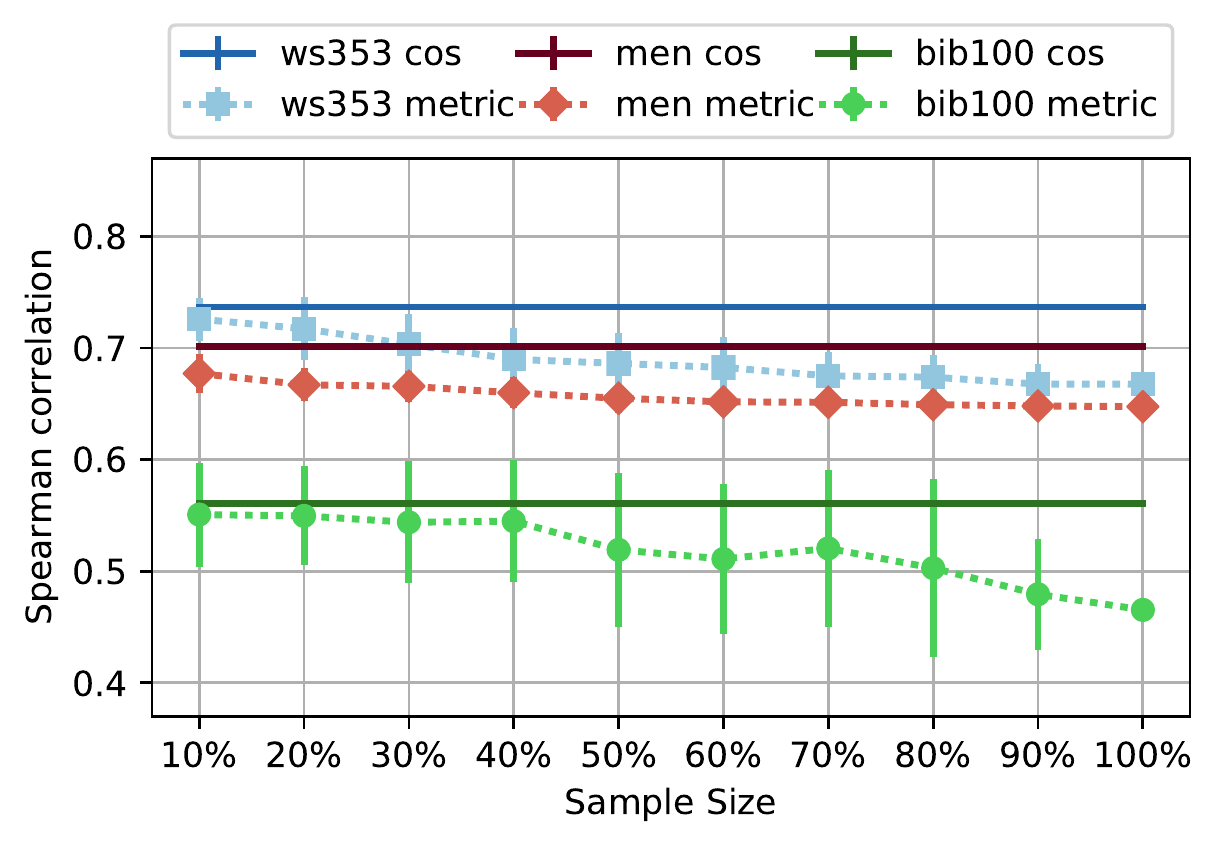}\label{fig:results_wikivectors_randomsampling}}
	
	\caption{Results for the robustness experiment. For each split, the
	scores of the word pairs in the training dataset were shuffled. 
	The test dataset stayed the same.}
	\label{fig:results_kaputtmaching}
\end{figure}
In \Cref{fig:results_kaputtmaching}, we can see that shuffled relatedness scores
exhibit a negative influence on the learned metric relatedness measure, as
expected.
On \bibs embeddings, only the measures trained on the \men dataset yielded
increasingly bad results, while the measures trained on the \bibeval or \wsbig
datasets did not change very much, though they did not improve correlation
either.
All measures trained on \delicious embeddings dropped in performance.
\wsbig-trained measures stayed bad and showed even worse results,
\bibeval-trained measures also decreased in performance.
Most notably, all embedding datasets except \bibs seem to be very receptive for
changes induced by \men relations:
While performance increased in the first experiment, here, it decreased
by a notable margin.
Nevertheless, the decrease of all tested measures is mitigated by the inherent
semantic content of the embeddings.
Overall, this shows the robustness as well as the consistency of our approach,
as was the goal of this experiment.

\section{Discussion of the Results}
\label{sec:discussion}

\para{Integrating Different Amounts of User Feedback.}
It can be seen that in the random sampling experiments, the
information in the \men dataset is generally best suited to learn semantic
relatedness.
While this might be due to its big size, it might also be due to this dataset
was created\furl{https://staff.fnwi.uva.nl/e.bruni/MEN}:
Using crowdsourcing, each human worker was shown two pairs of words and had to
determine which of both pairs is more related.
The higher a word pair in \men is rated, the more often it was considered more
related than the other one that was given.
Our approach exploits very similar constraints for learning.
Keeping this explanation in mind, we are able to give a recommendation on how to
gather human feedback in order to learn semantic relatedness with our method.
The bad performance of \men on the \bibs embeddings, both with the baseline and
the metric, could be attributed to the very low overlap of the \bibs
vocabulary and the \men pairs (see \Cref{tbl:hid_overview}) as well as the
small size of the \bibs tagging data.
On all other embedding datasets, where \men is very useful to learn the metric,
the pair overlap is notably higher.
Throughout all settings in this experiment, we observed that using more
data results in better relatedness scores, if there was a positive effect.

\para{Transporting User Intentions.}
The most notable result here is
that knowledge transfer from one HID to another works best and with large
improvements on the \wikiglove embeddings.
These embeddings are generated from the by far biggest word collections, \ie the
\wiki of 2014 as well as the Gigaword5
corpus.\furl{https://nlp.stanford.edu/projects/glove/}
It thus seems plausible that the embedding vectors encode a large portion of the
semantic information of the underlying corpus and thus benefit most from the
injected knowledge in our approach.
It also shows that we are able to adapt the metric and transfer the knowledge
if the vectors representations contain all necessary information.
\tni{Irgendwie gefällt mir der Satz so noch nicht.}
This is also in line with the notion that \bibs is the smallest corpus in our
collection, which seems too sparse to properly learn semantic relatedness.
\Cref{tbl:hid_results_bibs} shows almost only deteriorating correlation scores
when applying a learned relatedness measure, except for the \men-trained measure
evaluated on \wsbig.
Finally, results on \wikinav do not change very much, except when training the
measure on \men and evaluating it on \wsbig, which is a similar phenomenon as on
the \bibs embeddings.

\para{Robustness.}
On all four embedding datasets, evaluation performance decreases notably with
the \men dataset, with the worst performance loss on \bibs, where \men does not
perform well anyway.
We observe similar responses on \wikiglove.
These results confirm (again) that word embeddings successfully manage to
encode semantic information, and also that we cannot just ``unlearn'' it.
Furthermore, all embedding sets except \bibs react the most when used with a
measure trained on \men.
We attribute this to the same reasons as why \men is seemingly best suited to
learn semantic relations from, \ie it is constructed in a very similar way to
the form of the constraints that the learning algorithm is parameterized with.
Another consequence of this is that the promising results of the previous
experiments are indeed caused by the successful injection of semantic side
information into the relatedness measure.

\para{Additional remarks.} We are well aware that with our current set of
vector embeddings, we do not improve upon the current state-of-the-art evaluation
results on \wsbig and \men.
However, this was not our goal in this work, as we wanted to demonstrate the
feasibility of our metric learning approach to inject prior knowledge from human
feedback into a semantic relatedness measure.

\section{Related Work}
\label{sec:relwork}
In the following, we will report the most
relevant work in these fields of metric learning as well as semantic
relatedness learning algorithms.

\para{Metric Learning.}
Since we focus on the adaptation of metric learning on semantic relatedness
constraints, we give a short overview of different types of metric learning
algorithms.
Metric learning algorithms can be roughly split in two classes according
to the nature of the exploited constraints. 

The first class of metric learning algorithms utilizes link-based constraints,
\ie we have explicit information if two items are either similar or dissimilar.
As one of the first to propose an approach to learn a distance metric, Xing et
al.~\cite{xing2003distance} proposed to parameterize the Euclidean metric  with
a Mahalanobis matrix $M$ in order to improve kNN clusterings by incorporating
side knowledge.
Weinberger et al.~\cite{weinberger2009distance} presented the LMNN algorithm,
which aims to improve kNN clustering by placing items with similar classes near
to each other, while pushing away items with different classes by a large
margin.
The metric learning algorithm proposed in~\cite{davis2007informationtheoretic}
makes use of quadruplets $(x, x', y, y')$ and distance constraints $u$ and $l$.
These parameters can be translated to constraints $d(x, x') > u$ and $d(y, y') <
l$.
While the form of the constraints seems very similar to those of our approach,
the constraints still only encode two classes of similar and dissimilar items.
Finally, Qamar and Gaussier~\cite{qamar2009online} propose an algorithm to learn
a generalized cosine measure to improve kNN classification.
This approach is similar to ours, as we also learn a generalized cosine measure,
but their algorithm again exploits similarity and dissimilarity constraints
with a large margin.

The other class of metric learning algorithms is based on relative constraints,
\eg for three items $x, y, z$, a constraint could be $x$ is more similar to $y$
than $x$ to $z$.
This setting is much more in line with the idea of actually measuring a
continuous degree of relatedness instead of a preprocessing step for
classification or clustering.
In~\cite{schultz2004learning}, Schultz and Joachims propose an early distance
metric learning approach based on Ranking Support Vector Machines.
Their algorithm takes sets of triplets $(x_i, x_j, x_k)$ which encode the
constraints $d(x_i, x_j) < d(x_i, x_k)$, \ie $x_i$ is more similar to $x_j$ than
to $x_k$.
These constraints are extracted from clickthrough data, where explicit
preference information of a list of items compared to a reference item is
available.
This is however not the case in our scenario, as our constraints are based on
distance comparisons between four different items instead of only three.
The algorithm proposed in \cite{liu2012metric} makes use of relative distance
comparisons encoded in quadruplets $(x, x', y, y')$, which encode relative
comparisons $d(x, x') < d(y, y')$ without a separation margin, in order to learn
a metric.
This is a more general approach than the one provided
by~\cite{schultz2004learning}, as it is easy to convert triplet constraints to
quadruplet constraints, but not the other way round.

\para{Learning Semantic Relatedness.}
While the task to correctly determine the semantic relatedness of words or
texts has been around for a long time, there are still few approaches
which actually learn semantic relatedness.

Lately, many unsupervised approaches to learn semantic relatedness in low
dimensions have been proposed.
These methods are also often called \emph{word embedding algorithms}.
Such methods train a model to predict a word from a given
context~\cite{mikolov2013distributed,tang2015largescale,bengio2003neural,collobert2008unified}.
Other embedding methods focus on factorizing a term-document
matrix~\cite{deerwester1990indexing,pennington2014glove}.
These methods all have in common that they do not inject any external knowledge.
Anyhow, \cite{baroni2014count} showed that all those methods generally exhibit a
notably higher correlation with human intuition than the standard
high-dimensional vector representations proposed by~\cite{turney2010frequency}.

Bridging the gap between unsupervised relatedness learning approach and human
intuition by injecting side knowledge can be accomplished with post-processing
methods or with directly injecting this knowledge in the embedding process.
Both \cite{halawi2012largescale} and \cite{mrksic2016counterfitting} propose
approaches to inject synonymy and, in the case of the latter, also antonymy
constraints into semantic vector representations.
They aim to maximize the similarity of synonymous words, while minimizing the
similarity of antonyms.
Hereby, synonymy constraints acted as attractors in the semantic vector space,
while the antonymy constraints acted as repellants.
\cite{faruqui2014retrofitting} presented a method to fit the embedding vectors
to the neighborhood defined by relations in semantic lexicons.
In a way, also this algorithm is based on similarity constraints, as the
distance between similar vectors is minimized.
However, all of these works did not incorporate the actual degree of
relatedness into their approaches, which is what we do in this work.
There also exist methods which incorporate side knowledge directly into the
embedding process, \eg~\cite{bian2014knowledge,yu2014improving}.
However, our metric learning approach works on any already existing set of
vector embeddings instead of actually training new word embeddings from raw
data.

\section{Conclusion}
\label{sec:conclusion}
In this work, we presented an approach to learn semantic relatedness from human
intuition based on word embeddings.
Our approach is scalable and fast in terms of constraints and produces
significantly improved results compared to the widely used cosine measure, while
yielding competitive results on human evaluation datasets.
We argued for the use of word embeddings instead of high-dimensional vector
representations for tagging data due to an improvement in their semantic content
and their clear reduction of computational complexity when learning a metric.

Concretely, we could show that we can exploit semantic relatedness
information from HIDs to more realistically assess semantic relatedness,
regardless of the underlying embedding dataset.
Aditionally, we were able to encode and transfer knowledge from one HID to
another, sometimes with a very large increase of correlation with human
intuition.
When training a metric on false information to assess the robustness of our
approach, we argued that this actually supports our results, as the algorithm
yields negative results, as we expected.
Transferred to our previous positive results, we are indeed able to inject valid
knowledge into our relatedness measure to produce a better fit to human
intuition than only with word embeddings.

Future work includes the exploration of other graph embedding algorithms for
tagging data, the exploration of crowdsourcing strategies to best gather data
suitable for metric learning and further adaptation of metric learning
algorithms to specific properties of social tagging systems.

{\footnotesize \para{Acknowledgements.} This work has been partially
funded by the DFG grant ``Posts II'' and the BMBF funded junior research group
``CLiGS'' (grant identifier FKZ 01UG1408).}
\renewcommand*{\bibfont}{\footnotesize}
\vspace*{-0.7cm}
\printbibliography

\end{document}